\documentclass[conference]{IEEEtran}
\IEEEoverridecommandlockouts
\usepackage{amsmath,amssymb,amsfonts}
\usepackage{algorithmic}
\usepackage{textcomp}
\usepackage{xcolor}
\def\BibTeX{{\rm B\kern-.05em{\sc i\kern-.025em b}\kern-.08em
    T\kern-.1667em\lower.7ex\hbox{E}\kern-.125emX}}
    \usepackage{graphicx}
\graphicspath{{images/}}
\usepackage[style=ieee, citestyle=numeric-comp, backend=biber]{biblatex}
\begin{document}

\title{Deep Learning and Computer Vision Techniques for Microcirculation Analysis: A Review*\\
\thanks{Supported by the Norwegian Research Council and ODI Medical AS.}
}

\author{\IEEEauthorblockN{Maged Helmy}
\IEEEauthorblockA{\textit{Department of Informatics} \\
\textit{University of Oslo}\\
Oslo, Norway \\
magedaa@uio.no}
\and
\IEEEauthorblockN{Trung Tuyen Truong}
\IEEEauthorblockA{\textit{Department of Mathematics} \\
\textit{University of Oslo}\\
Oslo, Norway }
\and
\IEEEauthorblockN{Eric Jul}
\IEEEauthorblockA{\textit{Department of Informatics} \\
\textit{University of Oslo}\\
Oslo, Norway }
\and
\IEEEauthorblockN{Paulo Ferreira}
\IEEEauthorblockA{\textit{Department of Informatics} \\
\textit{University of Oslo}\\
Oslo, Norway }
}

\maketitle
\pagestyle{plain}

\begin{abstract}
 
The analysis of microcirculation images has the potential to reveal early signs of life-threatening diseases like sepsis. Quantifying the capillary density and the capillary distribution in microcirculation images can be used as a biological marker to assist critically ill patients. The quantification of these biological markers is labor-intensive, time-consuming, and subject to interobserver variability. Several computer vision techniques with varying performance can be used to automate the analysis of these microcirculation images in light of the stated challenges. In this paper, we present a survey of over 50 research papers and present the most relevant and promising computer vision algorithms to automate the analysis of microcirculation images. Furthermore, we present a survey of the methods currently used by other researchers to automate the analysis of microcirculation images. This survey is of high clinical relevance because it acts as a guidebook of techniques for other researchers to develop their microcirculation analysis systems and algorithms.

\end{abstract}


\section{Introduction} 
\label{intro}
One of the main functions of capillaries, the smallest vessels in the human body, is to deliver oxygen to the cells of all organs \cite{De_Backer2007}.
Networks of capillaries are known as the body’s microcirculation \cite{Guven2020-my}.
In the past decades, research has been conducted to elucidate whether monitoring microcirculation can be used to diagnose or assess the severity of various diseases \cite{arefa2020,Shore2000-mu}.
One of the first study directions was to record blood movement in the nail fold capillaries to assess the severity of rheumatic diseases such as systemic sclerosis, Raynaud’s syndrome, and dermatomyositis \cite{maricq1976skin}. These studies revealed that patients with rheumatic diseases had altered microcirculation, showing dilated and distorted capillary loops and areas with low capillary density compared with controls \cite{maricq1976skin}. More recently, studies on the effect of nonrheumatic diseases on microcirculation have been conducted. Microcirculation has been monitored either on the skin surface \cite{wester2014skin} or sublingually \cite{de2007evaluate} using video recordings of segments of blood flow within the capillaries. The impact of various diseases on microcirculation has been assessed, including sepsis  \cite{de2002microvascular}, \cite{top2011persistent}  and more recently, COVID \cite{natalello2021nailfold, edul2021microcirculation}. These studies have reported various correlations between diseases and the density of capillaries, the velocity of blood flow, and
 the heterogeneity of perfusion \cite{maricq1976skin,wester2014skin,de2007evaluate,natalello2021nailfold,edul2021microcirculation}.
 
Recently, the microvascular community has been focused on standardizing the analysis of microcirculation images \cite{ince2018second} by using automated methods \cite{hilty2019microtools,hilty2020automated,helmy2021capillarynet}.
This is of exceptionally high importance as, currently, the gold standard for microvascular analysis is manual analysis by a trained researcher \cite{hilty2019microtools}. This process is time-consuming (approximately 2 minutes per microcirculation image) and prone to subjective bias. Moreover, the length of analysis and requirements for trained researchers prevent microvascular monitoring from being used in routine clinical applications \cite{martini2020compelling}. In this article, we present the relevent methods in deep neural networks and traditional computer vision algorithms that can be used to achieve the automation of microcirculation image analysis.

Traditional computer vision techniques require the user to find the optimal set of values to segment the capillaries in an image, while deep neural networks can automatically attempt to find those values based on the dataset provided \cite{o2019deep}.
Though deep neural networks can automate the manual segmentation process, traditional computer vision algorithms require less computational power and are faster than deep neural network methods \cite{voulodimos2018deep}.
Thus, we present both techniques in this research paper.
That being said, we believe that a combination of both of these methods will be the future of microcirculation analysis in the clinical setting \cite{helmy2021capillarynet}.

This paper is intended to serve as a guidebook to inform researchers on what deep learning and computer vision techniques exist that can be used for the automated quantification of capillaries. The goal of the methods represented in this paper is to reduce the labor-intensive, time-consuming analysis from several minutes to seconds. Furthermore, these methods aim to reduce the subjectivity of the inter-observer variability using a standardized method.
Section \ref{capillary_data} presents and discusses the capillary dataset.
Section \ref{machine_learning} introduces machine learning and its four types.
Section \ref{DeepLearning_main} goes into the details of convolutional neural networks.
Section \ref{DeepLearning_OD} introduces the two object detection techniques: regional proposal based framework and unified based framework.
Section \ref{traditional} introduces traditional computer vision object detection techniques.
Section \ref{state_of_the_art} presents the current methods and techniques published and used by researchers for capillary quantification.
We then conclude our paper.

\section{Capillary Dataset} 
\label{capillary_data}

\begin{figure}[t]
\begin{center}
\includegraphics[width=\columnwidth]{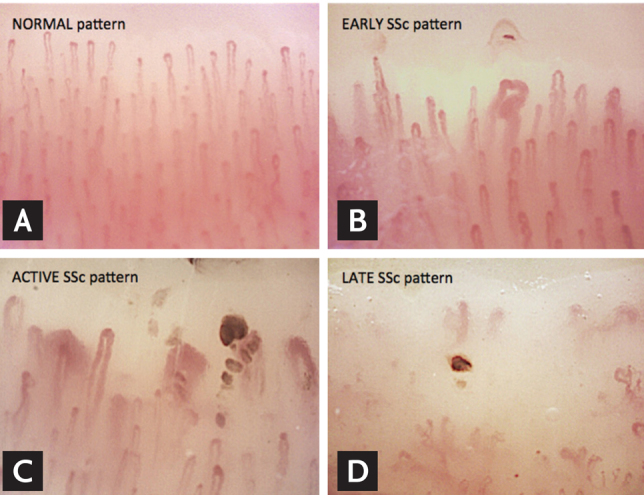}
\end{center}
\caption{Systemic sclerosis disease affects the microvascular structure. In image a, we see a normal pattern, while in images b, c, and d, we notice changes in the capillary structure as the disease advances. This image is from Ruaro et al \cite{ruaro2015methods}.}
\label{example} 
\end{figure}

Capillaroscopy is a method that noninvasively checks the dermal papillary capillaries using a microscopy system \cite{cutolo2008capillaroscopy}. The image(s) obtained can then be evaluated for capillary density, dimension, and morphology  \cite{cutolo2008capillaroscopy}. Figure~\ref{example} was obtained from Ruaro et al. ~\cite{ruaro2015methods} and best illustrates this. Ruaro et al. ~\cite{ruaro2015methods} describe systemic sclerosis as a disease that alters the microvascular structure, which can be seen using capillaroscopy. Figure \ref{example}a shows the typical pattern with no disease; Figures~\ref{example}b, c, and d show how the capillaries react to different disease stages~\cite{ruaro2015methods}.

\begin{figure*}[ht]
\begin{tabular}{cc}
\includegraphics[width=\columnwidth]{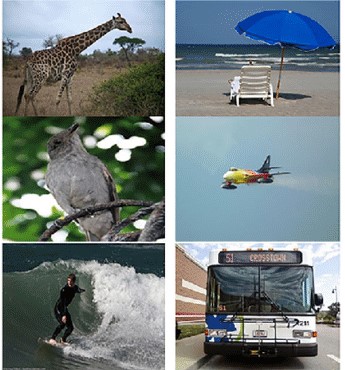}
&
\includegraphics[width=\columnwidth]{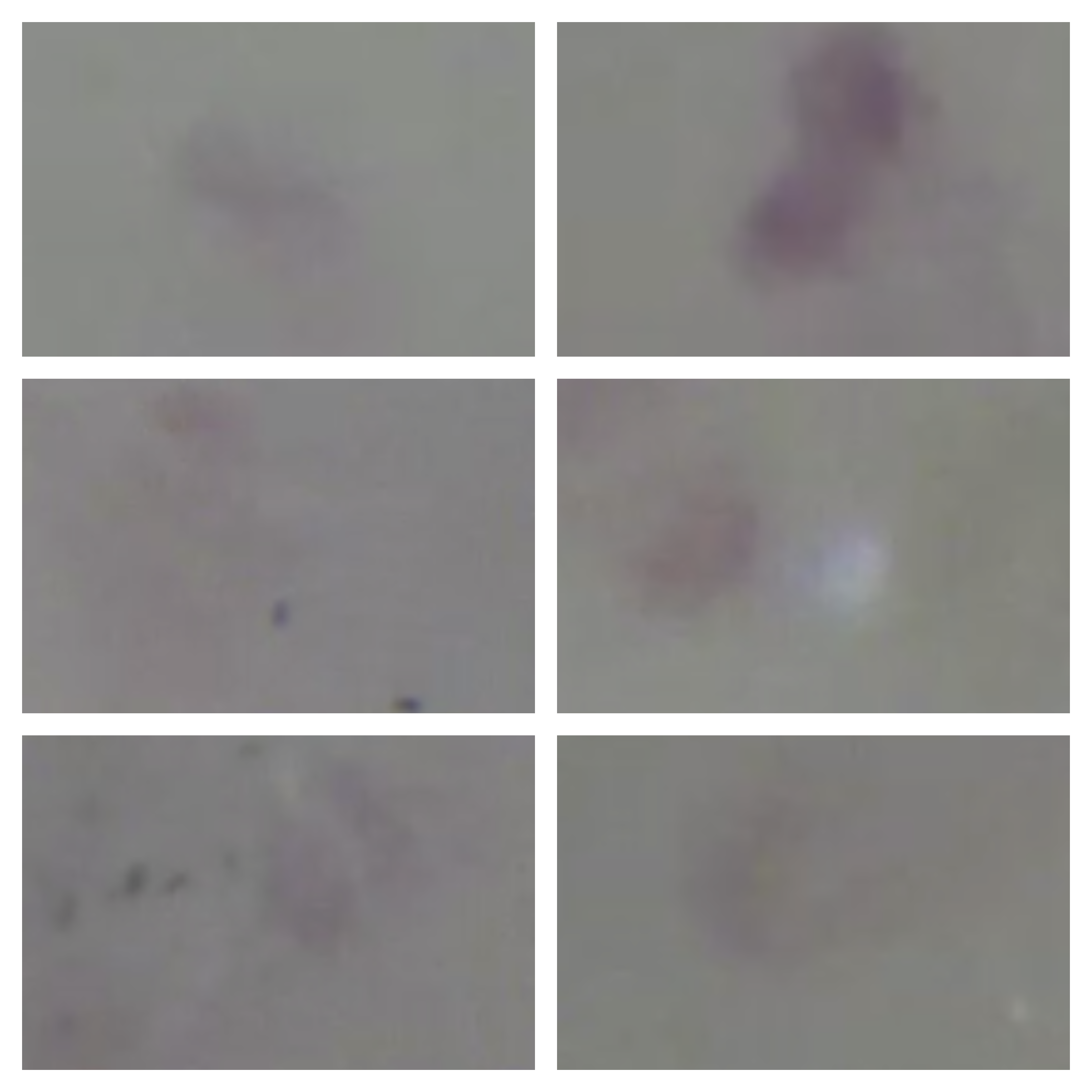}
\\
a & b
\end{tabular}
\caption{(a) Example of COCO \cite{russakovsky2015imagenet} set images, where the objects to be identified are clear with defined borders, and (b) Example of capillary dataset captured using a relatively high-end microscope. We see that the shapes of capillaries are not the same and can be faded. We also see instances with black spots and other cases with white spots, which is a reflection due to the oil. These are considered artifacts and are undesirable. Such issues make it challenging to develop a highly accurate algorithm that can generalize in detecting capillaries.}
\label{dataset_images} 
\end{figure*}

Several microscopes are currently available that can achieve such images: Dino-Lite CapillaryScope \cite{user}, Optilia Digital Capillaroscope \cite{optilia}, Inspectis Digital Capillaroscope \cite{inspectis_2020}, and Smart G-Scope \cite{scope_2021}. Other equipment that can capture capillaries include dermatoscope, ophthalmoscope, and stereomicroscope; however, none was designed for capillary capture. Therefore, images from this equipment are of relatively lower quality \cite{anders2001differentiation} than images captured by microscopes. That said, contemporary microscopes designed for capillaroscopy still do not produce adequate image quality compared to those in a standard object recognition dataset. Some examples of standard object recognition datasets used to develop object recognition algorithms are ImageNet (14 million+ images and 21,841 categories) [26], Common Objects in Context (COCO) (328,000+ images and 91 categories) \cite{russakovsky2015imagenet}, Places (10 million+ images and 434 categories) \cite{zhou2017places}, and Open Images (9 million+ and 6,000+ categories) \cite{kuznetsova2020open}. These datasets have several thousand images per category,  and the objects to be identified are relatively not as pixelated as the capillary data (see Figure~\ref{dataset_images}).Thus one of the main challenge is the dataset’s quality. Since the capillaries are relatively smaller in size, measuring less than 20 micrometers in diameter~\cite{De_Backer2007}, the shape of the capillary tends to be pixelated. Furthermore, in the literature, there is no clear definition of capillary shapes, which exponentially increases the challenge of not enough data being present for each shape/type of capillary, as shown in Figure figure~\ref{dataset_images}b.

\section{Machine Learning} 
\label{machine_learning}

Machine learning (ML) is a subbranch of AI that combines techniques from computer science, and statistics \cite{jordan2015machine}.
ML aims to find patterns from data to make predictions about new data \cite{zhang2020machine}.
This process of finding a pattern from a set of data is known as training \cite{burkov2019hundred}.
The product of this process is a model that is used to predict new data \cite{burkov2019hundred}.
An ML framework can consist of seven processes \cite{mayo2018frameworks,willemink2020preparing,ward2016general}, which are data collection, data preparation, selecting and training the model, testing the selected model, evaluating the model using F1-score, and finally deploying the model to production.

ML techniques can be further divided into supervised learning, unsupervised learning, reinforcement learning, and transfer learning \cite{kohli2017implementing}.

Supervised learning is an ML technique where the algorithm learns from data labeled by a human expert \cite{caruana2006empirical}. In terms of microcirculation data, the model will attempt to highlight the capillaries with a bounding box based on the sample of labeled data provided to it. Supervised learning can be further divided into regression and classification. Regression predicts a value, while classification predicts a class. In the case of microcirculation analysis, we predict whether a region has capillaries or not, so using the classification algorithms from the supervised learning techniques is the most suitable method for capillary detection.

Unsupervised learning, as opposed to supervised learning, is giving a set of data with no labels to the algorithm \cite{dayan1999unsupervised}.
The algorithm then attempts to find patterns between the data \cite{friedman2001elements}.
There are two types of unsupervised learning: clustering and association \cite{bousquet2011advanced}.
Clustering attempts to find subgroups in the data based on color, density, or other features. In contrast, association tries to discover relationships between patterns.
For example, a system that tries to recommend to the user what to purchase next based on what other users have bought, that is, association. Another example is anomaly detection, where the system finds data that exhibits patterns outside the normal range of the data. We did not find any suitable unsupervised algorithms for capillary detection. Moreover, we have not yet found papers on capillary detection that use unsupervised algorithms for capillary detection.

Reinforcement learning finds patterns based on a reward system \cite{sutton2018reinforcement}. When the system predicts the right output, the weights of those nodes in the neural network are increased, while an incorrect prediction diminishes the weights of other nodes \cite{szepesvari2010algorithms}. Reinforcement learning is common in game development for games like chess and Go \cite{wiering2012reinforcement, silver2018general}. We did not find any suitable unsupervised algorithms for capillary detection.

Transfer learning uses a model trained on previous data and tweaks the weights of the last n (defined by a machine learning expert) number of layers to train it on new data \cite{pan2009survey}.
An alternative way is to add new layers in addition to the existing layers of the network.
It is assumed that the earlier layers of a transfer learning model detects the generic features of the image like the width, height, and edges and that the latest layers learn the fine details of the classification \cite{weiss2016survey}.
Therefore by freezing the earlier layers while training the later layers on a new set of data, one can adapt any model to different datasets without having to retrain the algorithm from scratch \cite{torrey2009transfer}.

For microcirculation capillary quantification, our research work reveals that the most relevant ML techniques for microcirculation analysis are supervised classification learning and transfer learning.

\section{Deep Learning} 
\label{DeepLearning_main}

Deep learning is a sub-branch of ML that has vastly advanced the state of the art in speech recognition and visual object detection \cite{lecun2015deep}.
In its simplest form, a deep neural network is composed of multiple layers of neurons \cite{yan2015deep}. This typically consists of the input layer, hidden layer(s), and the output layer \cite{lecun2015deep}. Deep learning outshines the other ML techniques because it can construct a feature extractor without the need for domain expertise \cite{goodfellow2016deep}. Rather, the neural network increases prediction accuracy by adjusting the neuron weights to optimize an appropriate cost function using  back-propagation techniques \cite{shen2017deep}.
Deep learning architectures combine multiple non-linear modules to transform input data into a higher abstract level.
That suppresses irrelevant information while intensifying the useful parts to increase prediction accuracy \cite{deng2014deep}.
In this section, we describe several deep learning architectures and then delve into the details of convolutional neural network (CNN) architectures.

\subsection{Types of Deep Neural Networks:} 
\label{deep_neural_network_types}

There are many types of deep neural networks. In this section, we will focus on the three most relevant for microcirculation analysis: recurrent neural networks (RNNs) \cite{medsker2001recurrent}, generative adversarial networks (GANs) \cite{creswell2018generative}, and CNNs \cite{o2015introduction}.

RNNs are a type of neural network that has a hidden state to find patterns in sequential data and uses the output of previous layers as an input for the current layer \cite{medsker2001recurrent}.
The nodes of an RNN are connected by directed cycles, allowing it to keep the state of the previous nodes. While a traditional neural network takes in the input and gives an output, an RNN assumes a relationship exists among the sequences of input data \cite{mandic2001recurrent}. Thus, RNN is mainly used in speech recognition, translation, and sequential data analysis \cite{mandic2001recurrent}.

Long short-term memory networks (LSTM) \cite{cheng2016long} and gated recurrent units (GRU) \cite{dey2017gate} are RNN architectures that deals with gradient vanishing gradient challenges \cite{chung2014empirical}.
The vanishing/exploding gradient problem is when the size of the gradients of each layer does not equal 1.
This will prevent the neural weights from converging.
The modifications introduced by LSTM and GRU deal with this challenge by reducing the amount of irrelevant information propagating through the architecture by adding appropriate gates into the vanilla RNN.

Generative Adversarial Networks (GAN) creates new instances of the data \cite{pan2019recent}.
They are made up of a generator (G) and a discriminator (D), whose tasks are opposite to each other.
The generator obtains a random noise input z, and its output is a fake datum sample G(z).
The discriminator D obtains as input both G(z) and a real sample x, and its output is the probability of whether G(z) is real.
G can reliably produce new instances that mimic those in the real dataset.
% One of the main uses of GAN is in data augmentation to create more artificial data resembling the real data as much as possible. Furthermore, they can be used to enhance the image by scaling it up and increasing its resolution \cite{saxena2021generative}.
% The generator creates a fixed length of random values drawn from a probability distribution that closely represents the values of the input data. The values of this probability distribution is dictated by the discriminator which uses the input data to generate these values. For example, everytime the generator generates an image, the discriminator classifies it if its resembles the real image or not. If not, the generator draws on a new set of random images, until the discriminator classifies it as a real image. All the real images accepted by the discriminator then form the augmented dataset.
One of its main uses is in data augmentation to create more artificial data resembling the real data as much as possible. Furthermore, they can be used to enhance the image by scaling up the image and increasing its resolution \cite{saxena2021generative}.

CNNs are the most prominent architecture used for medical image analysis, specifically for microcirculation analysis  \cite{shen2017deep}, \cite{anwar2018medical,litjens2017survey,yamashita2018convolutional}.
They were originally developed in 1998 to recognize zip codes \cite{lecun1998gradient} and digits.
Gradually, they have become the most relevant architecture in image classification.

\subsection{Convolutional Neural Networks} 
\label{Types_CNN}

In this subsection, we will dive into the details of a CNN.

\subsubsection{A CNN Pipeline} 
\label{cnn_pipeline}

A typical CNN image classification pipeline can consist of three main stages, which are:

\begin{itemize}
\item Dataset: Where a set of images is labeled with their corresponding classes,

\item Learning: A model that has learned from every class of the data,

\item Evaluation: Evaluating the performance of the model on a set of data that it has not been previously trained on. \cite{el2019deep}. 

\end{itemize}

The data at the first step can be further split into a training set and a validation set, so one can evaluate the model before new data comes in.
In this way, a pipeline can end up with three sub-datasets: a training set, a validation set, and a test set. 
A good CNN should be invariant to the listed challenges below \cite{zhao2019object}:

\begin{itemize}
\item  Variation of viewpoint: As regards the camera, a single instance of an entity may be directed in various ways,

\item Variation of scale: The scale of visual groups is often variable,

\item Deformation: A term used to describe the process of changing many interesting things, which are not solid bodies and may be deformed dramatically,

\item Occlusion: A term used to describe a situation where objects of interest can be obscured. Just a small portion of an item (a few pixels) can be observable at any given moment,

\item Illumination: Different brightness levels on different parts of the image can significantly affect algorithm performance,

\item Clutter in the background: The objects of interest can blend in with their surroundings, making them difficult to spot,

\item Variation within a class: The categories of interest, such as chair, may be very general. These artifacts come in various shapes and sizes, each with their distinct appearance.

\end{itemize}

Data preprocessing can be done in two ways: mean subtraction and normalization \cite{karpathy2018stanford}.
The most popular form of preprocessing is mean subtraction \cite{karpathy2018stanford}.
It entails subtracting the mean from all the data’s features, with the geometric understanding of centering the cloud of data (pixel values) in all dimensions (1 in case of white and black, 3 in case of RGB).
Normalization is the process of bringing the sizes of the scale closer together.
This normalization can be accomplished in two forms.
The first is to zero-center the value and divide it by its standard deviation.
The second is to normalize each dimension such that the minimum and maximum values are between negative 1 and positive 1.
In the next section, we will talk about the different parts that make up a CNN: the fully connected layer, the convolutional layer, and the pooling layer \cite{o2015introduction}.

\subsubsection{The Anatomy of a  fully connected layer} 
\label{DeepLearning_anatomy}

In this subsection, we will describe the details of a fully connected layer \cite{karpathy2018stanford,yegnanarayana2009artificial,yiqiao2018deep}.

Weight initialization: Each neural network node consists of parameters known as weights, which take numerical values and contribute to the total of the other weighted input signals.
It is recommended to initialize a neural model’s weights with a positive random value to help it converge into the results with the epochs. There are other ways to initiate the weights in a neural network using Xavier weight initialization and He weight initialization. At every iteration, the algorithm tries to find the most appropriate set of weights for each node. For example, in stochastic descent, the neural network weight is gradually adjusted by a cost function.

A cost function is used in supervised learning to measure the difference between the predicted result and the expected value. There are two types of loss functions: One is used for classification when the classes are of a fixed size and set, while regression is used to predict a quantified value.

Regularization: There are four ways to regularize a neural network: L2 regularization, L1 regularization, max norm constraints, and dropout. The most popular form of regularization is L2 regularization. It can be applied by penalizing all parameters’ squared magnitudes. The intuitive understanding of the L2 regularization is that it strongly penalizes high weight vectors while favoring diffuse weight vectors. In L1 regularization, the values become almost invariant to noisy values since they use the most critical inputs and almost all of their irrelevant information is eliminated. In general, if a CNN architecture would be used for feature selection, then L1 regularization handles it the best over other regularization methods.

Maxout constraints on the maximum average, which is another method to regularize the values by imposing an absolute upper limit on the magnitude of each neuron’s weight vector, is enforced using projected gradient descent. In fact, this entails updating the parameters as usual and then clamping the weight vector to enforce the constraint. Dropout is an easy, efficient tool for removing a random percentage of neurons that the developer of the neuron can specify.

Activation Function: This part in a CNN takes in a number and performs a mathematical operation on it. Many activation functions can be performed in a CNN; however, the common ones are sigmoid, tanh, ReLU, leaky ReLU, and maxout.

Hyper-parameter optimization is finding the right set of values for all of the above properties in a neural network. Mainly, finding the initial learning rate, the decay constant of the learning rate, and the regularization values can strengthen or weaken the neural network.
During the forward pass, the score is calculated by applying the operations of all the blocks on the input value. On the backward pass, we compute an updated value of the weights that minimizes the loss function, which increases the overall accuracy of prediction.

\subsubsection{The Convolutional layer and the Pooling Layer} 
\label{CNN_pooling}

In this section, we will describe the details of the convolutional and pooling layer \cite{zhao2019object, karpathy2018stanford, kim2017convolutional}.

Similar to the anatomy of a node in a fully connected layer, explained in the previous section, a convolutional block is initialized with weights and has a cost function, activation function, and regularization. The difference is that convolutional net architectures presume that inputs are pictures, taking the spatial variance between the inputs into consideration.
These processes and parameters decrease the network’s overhead, reducing the number of parameters the program will execute as well as increasing the speed. If we were to pass a full HD image to a neural network (1920x1080x3), there would be approximately 6.2 million weights to initialize—one weight for each pixel. Instead, we pass that image to a series of convolutional blocks first to reduce the number of weights from 6.2 million to possibly several hundred thousand weights, without compromising accuracy. This reduction of pixels is achieved by transforming the image into a more representative form using a CNN architecture.

A simple CNN architecture consists of four parts: 
\begin{itemize}

    \item 
    input, where data is loaded as a matrix in its raw form (for example, a full HD image will be a matrix of 1920x1080x3),
    
    \item 
    a convolutional layer, which takes in a value and applies the dot product operation followed by some kind of non-linearity (activation function),
    
    \item 
    the pooling layer, which applies downsampling to the matrix; and
    
    \item 
    the Fully Connected (FC) layer, which computes a prediction associated with each class. With each iteration, the values for the convolutional and FC layers change as their weights are adjusted, while the activation function and pooling layer values stay constant throughout the whole process. The details of each part are described below.

\end{itemize}

A convolutional layer consists of a set of filters, forming a matrix that is typically 5x5x3. However, these values are strictly experimental and depend on the image used. A convolutional layer can have any number of filters specified by the machine learning expert. The filter values are randomly generated and are updated in the back-propagation to reduce the loss value, which increases the probability of correct classification. Each filter slides over the image with a stride value production of a smaller image known as the activation map. The number of strides dictates the number of weights to be initialized later: as the number of strides increases, the number of weights to be initialized at the FC  decreases. For example, a stride of two means the filter will jump two pixels in the image before applying the dot product and skipping some pixels. There is a trade-off. Each filter can detect a different image property; for example, a filter can detect horizontal or vertical edges and types of colors. The number of filters in this layer is referred to as the filter depth. Each filter produces an activation map, and these are then stacked on top of each other. Padding is another concept in the convolutional layer that involves adding zeros around the borders of the input image to preserve the sizes of the input and output shapes. Combining the filter depth, stride, and padding, the output volume of the convolutional layer can be calculated.
Output Volume: (W-K + 2P / S) + 1, where W is the input height/length, K is the filter size, P is the padding, and S is the stride. 
For example, a single black-and-white image of dimension 200 with a stride of 1, padding of 0, and filter size of 5, with 32 filters, will turn the image from a 200x200x1 to a 196x196x32. Each layer represents the activation map produced by the individual filter. This number is very high compared to the input value, and thus a pooling layer is applied to take down the number of parameters to initiate weights for. A pooling layer is typically inserted between consecutive convolutional layers before passing the final value into an FC layer.
The pooling layer is applied individually on the activation map; the depth does not change, but the activation map's width and height are reduced. The pooling layer scales the image and takes in two parameters: the window size and stride. The larger these values are, the smaller the output image will be. There are several types of pooling, but the most common are average and max pooling. Average pooling takes the mean of a window of pixels, while max pooling takes the maximum value within the window. So a pooling layer with a 2x2 window and a stride of 1 halves the image's dimensions from 196x196x32 to 98x98x2. Applying several of these between consecutive convolutional layers inevitably reduces the dimensions of the activation map to the most relevant for prediction.
The values from the last pooling layer are flattened or converted from 2D to 1D and passed to the fully connected layer explained earlier.
In the next section, we look into how the convolutional layers, activation function and pooling function are used as fundamental building blocks to predict the image classification.

\subsection{Types of Convolutional Neural Networks} 
\label{DeepLearning_CNN}

LeNet \cite{lecun1998gradient}, created in 1998, uses a five-level CNN: two convolutional layers with three fully connected layers. The convoluted layers were made up of a 5x5 filter of a stride of 2 with sigmoid function, followed by an average pooling layer of 2x2 with a stride of 1. The fully connected layer contained 120, 84, and 10 neurons, respectively, using softmax as an activation. This CNN's input data was a grayscale 32x32, which is relatively small with today’s standard.

AlexNet \cite{krizhevsky2012imagenet}, released in 2012, outperformed LeNet. It used an eight-layer-deep CNN: five convolutional layers, two hidden layers, and one fully connected output layer. There are several significant differences between AlexNet and LeNet. AlexNet uses ReLU for the activation function. Moreover, AlexNet uses dropout instead of weight decay for regularization. AlexNet uses more neurons and different filter sizes for each convolutional net. For the FC layer,  AlexNet uses 4,096, 4,096, and 1,000 neurons, respectively, compared to the 120, 84, and 10 of LeNet. AlexNet uses 11x11, 5x5, and 3x3 for the convolutional layer, while LeNet uses two 5x5 filters.

Visual Geometry Group (VGGNet) \cite{simonyan2014very}, developed in 2014 at Oxford University, consisted of these basic CNN building blocks: a convolutional layer, activation function ReLU, and a maximum pooling layer. It used 3x3 filters with a padding of 1 and a 2x2 pooling with a stride of 2. Moreover, the paper authors experimented with several different architectures and concluded that deeper and narrower layers get better results than fewer and wider convolutional layers.

The above three architectures have a common pattern of using the convolutional layer followed by pooling with minimal tweaks. The next three architectures came later and use a slightly different design pattern.

GoogLeNet \cite{szegedy2015going}, published in 2015, outperformed the previous three architectures, achieving close to human-level performance with its new inception module. An inception block uses four different blocks in the input images and then concatenates their output. The first three blocks apply a convolutional layer of different window sizes, while the fourth applies max-pooling and then the convolutional layer. The number of blocks in an inception module can be tested on different sets using hyperparameter tuning. GoogLeNet outperformed the others because it aims to extract the most spatial information possible from each layer. Instead of finding information from the previous block’s output, inception aims to explore each image with different filter sizes. It is like taking the same photograph with different lens magnifications.

Residual Neural Network (ResNet) \cite{he2016deep} was published in 2016 and designed to address the increasing complexity of making deeper neural networks. As the number of blocks increased, the accuracy gain per block decreased to where making the network even deeper started adversely increase complexity and computational power and reduce accuracy. This was achieved using the residual block, which utilizes a skip connection with heavy batch normalization. Like the VGGnet convolutional layer design, a ResNet consists of two consecutive 3x3 convolutional layers with an activation function. In addition, there is a connection between this block’s input and the output, known as the skip connection. Like GoogLeNet, which uses four modules within the inception block, ResNet uses four modules within the residual block. It also uses a global average pooling layer before the FC layer.

For microcirculation image analysis, it is not enough to use a CNN architecture; these architectures only detect whether an image has a capillary. They cannot pinpoint the capillaries’ location. To do that, we need to extend the CNN architecture with an object detection architectures. 
%Object detection architectures combine a CNN architecture with other methods to localize the location of the capillaries. The main difference between image classification and object detection is an image classifier consists of two modules: the feature extraction module and the classification module. Such image classifiers Include ResNet, VGG, Inception, DenseNet, etc. Object detection, on the other hand, is more architecture with several combinations of modules. The architecture may consist of a feature extraction module, a region proposal network, a regression model, U-Net, and many other modules.
%The image classification feature extraction is used as part of object detection. The image classification is sometimes referred to as the ‘backbone network’ to the object detection architectures.

\section{Deep Learning Object Detection Techniques} 
\label{DeepLearning_OD}

Object detection techniques aims to estimate the location and label of an object in an image \cite{amit2020object}.
The object detection part extends the CNN architecture.
Object detection techniques can generally be split into two distinct categories.
The first category, which is a two-step method, aims to first locate the object in the image (object localization) and then estimate the category of the object (object classification).
These architectures can be referred to as the Region Proposal Based Framework.
The second category is a one-step method which aims to locate and categorize the objects in one go.
These architectures are known as the Unified Framework \cite{liu2020deep}.
Before these methods were developed, the field was dominated by different techniques known as the scale-invariant feature transform (SIFT) technique from 1999 to 2012 \cite{lowe1999object,krizhevsky2012imagenet, alom2018history}.
Object detection architectures are benchmarked by measuring the mean Average Precision (mAP) and efficiency (speed of detection per frame) on standardized datasets \cite{everingham2010pascal, russakovsky2015imagenet,lin2014microsoft,zhou2017places, kuznetsova2020open,everingham2010pascal,hoiem2012diagnosing}.
In this section, seven selected architectures from the Region Proposal Based Framework are described along with five selected architectures from the Unified Framework.

% Object localization can be further subdivided into three categories: generic object detection, which is a bounding box around the location of an object \cite{everingham2010pascal}; semantic segmentation, which is a pixel-wise segmentation mask around the object(s) from the same class \cite{russell2008labelme}; and instance segmentation, which is a pixel-wise segmentation mask around an individual object regardless of its class \cite{lin2014microsoft}.
% Object classification can be further subdivided into two categories: object matching, which detects specific instances of the object localized (e.g., Oslo Town Hall, Norwegian Parliament), and generic object detection, which detects the category of the objects localized (e.g., skyscraper, dog, cat) \cite{zhang2013object}. The latter is more challenging due to the large variations in appearance differences within the same category—for example, the amount of light on the image, the object positioning, the rotation angle, if the object is mirrored, occlusion, distance from the camera, resolution, blur, and motion \cite{khurana2013techniques}. Furthermore, existing object detection architectures have been tweaked for domain-specific object detection cases like pedestrian detection \cite{ouyang2013joint}, face detection \cite{peng2016graphical}, salient object detection \cite{borji2015salient}, medical image analysis \cite{shen2017deep}, vehicle detection \cite{chen2014vehicle}, and text detection \cite{he2016text}. However, in this section, we will focus on the generic object detection architecture of those techniques.

\subsection{Region Proposal Based Framework}

R-CNN (Rich feature hierarchies for accurate object detection and semantic segmentation)  \cite{girshick2014rich}: When this paper was released in 2014, the best-performing object detection architectures had plateaued from 2010 to 2012, with an accuracy of 35\% for the most popular datasets \cite{alom2018history}. This algorithm achieved 20\% higher accuracy than its predecessor with the VOC 2012 dataset \cite{girshick2014rich}. This method was termed Regions with CNN features, or R-CNN. It is also one of the first methods that propose a two-step approach, and many subsequent methods have been based on this approach from 2014 to this day \cite{zhao2019object}. Region proposal based frameworks are inspired by the combination of Deep Convolutional Neural Network (DCNN) \cite{krizhevsky2012imagenet} and region proposals \cite{uijlings2013selective}. R-CNN takes in an image, applies a segmentation mask to it, and extracts the top 2,000 promising bounding boxes on that segmentation. The bounding boxes are of different scales, increasing the probability of identifying different sizes or shapes. It then computes the features of these boxes using a CNN and classifies each region with a linear SVM. On the other hand, this method can be slightly adjusted if the training data is low to apply a supervised pre-training CNN on the ImageNet followed by fine-tuning the low training data. The R-CNN consists of two concurrent pipelines: the region detector and the object detector. For the region detector, R-CNN takes an image and applies a non-deep learning model called selective search to extract approximately 2,000 regions of interest (RoI). These regions present the places in the image where an object is more likely to reside. The proposed region is then warped or cropped to fit a specific dimension before being passed into the object detector. The object detector applies CNN + max-pooling with an activation function to calculate the feature map. The feature map is then passed to an FC layer to create a 4096-dimensional vector. This vector is passed to a classification head that tries to figure out the class, and the regression network that tries to refine the box coordinates. The classification head is optimized using cross-entropy loss, while the regression head is refined using L2 loss. The model is trained by optimizing the model first on the classification loss, then the regression loss. This can take up to several days, with large storage space, since all the features computed from the proposed regions require many gigabytes. This paper methodology consists of three modules.

\begin{itemize}

\item Generating the region proposals: Selective search method is used to suggest the regions,

\item Extracting the features using CNN: A 4096-dimensional vector is extracted from each region generated by the previous step using the method applied by the DCNN [61]. The features are then computed by subtracting the mean from a 227x227 image through a five convolutional layer and a two-FC layer. The output region is warped equally with p=16; however, the paper suggests that alternative values can be used.

\item Extracting the class using SVM: Each region is scored using an SVM, and a greedy non-maximum suppression for each class is applied independently in the proposed region. If the two regions have an intersection-over-union (IoU) higher than a threshold, one region will be rejected,

\end{itemize}

If there is a lack of training sets, the paper suggests the addition of these two modules:

\begin{itemize}
\item Supervised pre-training: It starts by training the CNN on one of the large datasets as an image classification problem using the Caffe CNN library,

\item Domain-specific fine-tuning: The wrapped regions created in step two of the above method are used to fine-tune the CNN parameters using stochastic gradient descent.
\end{itemize}

R-CNN applies the results on the PASCAL VOC 2007 dataset and achieves a 53.7\% mAP.
These results are a big jump from the previous algorithms proposed for this dataset at that time (year 2010) where the highest achieved mAP was 35.1\%.
R-CNN is a big step toward building a high-quality object detection architecture after the SIFT era and in the DCNN era. This is noticeable in the jump in accuracy introduced by R-CNN. However, there are some drawbacks to using R-CNN. First, R-CNN has multi-stage, multi-step modules that need to be optimized individually to achieve good results, which increases the chances of introducing inaccuracies and makes training time notably longer. Second, R-CNN uses a fully connected layer that requires a fixed input shape.
Moreover, approximately 2,000 regions are extracted, which one can argue is way too much in sparse images and way too little in denser images. Such disadvantages have led to the development of successors such as SPP-Net (Spatial Pyramid Pooling), Fast R-CNN, Faster R-CNN, Region-based Fully Convolutional Network (R-FCN), Feature Pyramid Network (FPN), and Mask R-CNN which are presented in the next paragraphs.

SPP-Net (Spatial Pyramid Pooling in Deep Convolutional Networks for Visual Recognition) \cite{he2015spatial}: This method introduces two changes to the existing R-CNN architecture. First, it aims to tackle the challenge of having a fixed-size window since important information can be lost or distorted, reducing accuracy. Second, SPP-Net computes the feature maps for the images instead of repeatedly computing them on each ROI region as the R-CNN.
The challenge of having a fixed-size window is tackled by adding a spatial pyramid to the top of the last convolutional layer before the FC layer.
Instead of cropping or warping the image, this method aggregates the information by pooling the features and feeding it to the FC layer.
The spatial pyramid pooling is an extension of the Bag-of-Words model released in 2006 \cite{lazebnik2006beyond}.
The difference between the R-CNN method and the spatial pyramid pooling methods can be illustrated as follows: The R-CNN method takes in an image, applies crop/warp, and passes it to the convolutional layer and then the FC layer.
The SPP-Net method takes an image, passes it directly to the convolutional layer, applies the spatial pyramid pooling, and then passes it to the FC layer.
The SPP-Net takes in the feature maps from the last layer of the convolutional layer to create feature maps of fixed-length feature vectors regardless of the input image size.
%To summarize, SPP-Net calculates a feature map for the whole image and then classifies the regions using a feature vector. The features are then extracted using max-pool.
Images with different sizes can be pooled and aggregated into a spatial pyramid, which is then passed to the FC layer.
When this paper was released, four existing object detection architectures were compared with their non-SPP counterparts (ZF-5, Convnet*5, Overfeat-5, Overfeat-7), and the CNN with SPP-Net showed state-of-the-art classification results on Pascal VOC 2007 and ranked at number 2 on the ILSVRC 2014 competition \cite{zhao2019object}.
The spatial pyramid pooling method is more efficient than its predecessor since it obtains a significant speedup.
The speedup is due to the fact that the CNN layer generates a feature map by running one iteration on the image.
Furthermore, it is more accurate since it can learn feature maps from any scale without losing information to cropping or warping.
The multi-level pooling makes the input images more robust to deformation. The main drawbacks of SPP-Net are that it is still a multi-stage, multi-step pipeline (feature extraction, network fine-tuning, SVM training, bounding box regressor, feature caching), making it relatively slow. Furthermore, the authors of the paper mention that the accuracy of SPP-Net layers drops when using deeper CNN since tuning the network will not update the layers before the pyramid layer, leading to reduced accuracy and a very difficult challenge in implementing back-propagation.

Fast R-CNN \cite{girshick2015fast}: This paper addresses the problems arising from the SPP-Net and R-CNN architectures. Until this paper’s 2015 publication date, object detection methods required generating several hundred regions known as ‘proposals’ to create a feature map; then, the proposals generated estimated the localization of the object.
These proposals reduced speed and accuracy while increasing complexity. Similiar to R-CNN, this method uses selective search to find the regions and then passes the regions to the object detector. The method also consists of two SVM heads: one for classification to get the class category and the second regression to calculate the bounding box coordinates. The difference is that instead of running the CNN several times on the RoI, it runs it only once by introducing RoI pooling. Second, it streamlines the process on the object detector side, where it jointly classifies and learns the object’s location simultaneously by using multiclass loss. This method generally achieves a higher mAP by having a single stage for the training with a multi-task loss. The increased accuracy is obtained by updating all layers. The speed is achieved by not requiring the features to be cached and because Fast R-CNN learns the softmax classifier and bounding box regression together rather than in two separate processes. These improvements have led to a major decrease in storage space needed. Unlike the R-CNN, Fast R-CNN creates a feature map from the entire image.
%This method’s feature map is produced by applying convolutional neural networks and max-pooling to the input image. The regions are then cropped using RoI pooling from the feature map instead of the input image. The RoI pooling layer converts the proposals’  features into a feature map. An RoI is a rectangular window that is similar (one pyramid level) but not identical to the spatial pyramid pooling layer used in SPP-Nets. The cropping on the feature map is based on using selective search (RoI). The cropped regions are passed to a CNN where the output is a fixed-length feature vector 4078 dimensional. These vectors are individually fed into the FC layer, giving two outputs: 1) softmax probability estimates over the classes and 2) four real values that encode the bounding box position for one of the classes.
Furthermore, Fast R-CNN method improves efficiency compared to the SPP-Net: 3x in training and 10x in testing. The authors report that ‘Fast R-CNN trains 9x faster than R-CNN on the VGG-16 and 213x faster at test-time, with a higher mAP on the PASCAL VOC 2007, 2010 and 2012 dataset...’ [94]. These speed improvements result from a single process that updates all layers without requiring feature caching.
%One important improvement that this method introduces is the increase in back-propagation efficiency. Previously, R-CNN and SPP-Net were inefficient since the input to the network layer came from different images, and the region of interest might span the whole image. This method allows the computational sharing of RoIs from the same image since regions are applied on the feature map instead of the full image. 
Moreover, to reduce the time spent on the FC layers, Fast R-CNN uses a truncated singular value decomposition (SVD) to accelerate the testing procedures [106]. This method has significantly increased the speed and efficiency of object detection, firstly, by streamlining the whole process and, secondly, by applying SVD on the testing set. Thus, Fast R-CNN is more of a speed improvement than an accuracy improvement. On the same dataset that took 84 hours to train, Fast R-CNN performed it in 9 hours. A major drawback is that Fast R-CNN still relies on external region proposals that make the whole process relatively slow. It uses the selective search method to find the RoI. Furthermore, later research has concluded that convolutional layers are sufficient to localize objects; therefore, adding an FC layer slows down the process unnecessarily.

Faster R-CNN - Toward real-time object detection with region proposal networks \cite{ren2015faster}: Optimizations introduced by SPP-Net and Fast R-CNN have exposed the fact that using external region proposal methods slows down the process. Previous networks mainly relied on selective search (SS) \cite{uijlings2013selective} and Edge box \cite{zitnick2014edge} to create region proposals. This paper introduces a region proposal network (RPN), which aims to replace the SS and Edge Box. The RPN introduces an almost cost-free proposal computation. For RPN to compete with methods like selective search, it has to predict the RoI of multiple scales and ratios from an image much faster. Thus, RPN introduces a novel concept of creating anchors on the feature maps. The RPN layer takes in the feature map and generates rectangular object bounds using CNN, which are the new ‘regions of interest’. Faster R-CNN can be trained end-to-end like Fast R-CNN. The RPN output tells the Fast R-CNN where to look. The Faster R-CNN architecture is complex because it has several interconnected parts. The RPN first initializes anchors of different ratios and scales on the feature maps created by the convolutional layer. The paper’s author uses nine types of anchors when the RoI is decided on. The anchors are off three scales and three ratios. These anchors are mapped and fed into the two FC layers, where one layer is responsible for the category classification and the other for the box regression. RPN shares the convolutional feature with the Fast R-CNN, enabling the same efficient computation as mentioned in the methodology of the previous paper. On the VGG-16 model \cite{willemink2020preparing}, the Faster R-CNN efficiently performs all steps on a 5fps with an accuracy exceeding all recorded results on the VOC 07 dataset with 73.2\% mAP, and on VOC 12 with 70.4\% mAP.
Although this method is several times faster than Fast R-CNN, it still relies on applying several hundred RoIs per image to detect the region of interest. This leads to computations not being shared after the RoI layer, reducing this method’s overall efficiency.

R-FCN - Object Detection via Region-based Fully Convolutional Networks \cite{dai2016r}:
% The processes of using RoI pooling in Faster R-CNN meant that knowledge learned from each RoI detected were not shareable.
% Thus, the RoI pooling calculation had to be reapplied for each RoI.
In Faster R-CNN, each region proposal had to be cropped and resized to be fed into the Fast R-CNN network.
The R-FCN attempts to speed up the network by converting this process into fully convolutional.
It aims to swap the costly per-region subnetworks with a fully convolutional one, thus allowing the computation to be shared across the whole image.
Furthermore, the R-FCN differs from the Faster R-CNN in the RoI pooling layer.
The R-FCN proposes a method to use convolutional layers to create an RoI subnetwork.
It uses the RPN introduced in the previous method to extract features and pass them on to the R-FCN.
The R-FCN then aggregates the output of the last convolutional layer and generates the scores for each RoI.
Instead of cropping the regions from the feature map, the R-FCN inputs the feature map into the regression and classification heads, creating an RoI map on the feature map.
R-FCN uses ResNet-101 as the backbone of its architecture \cite{he2016deep}.
ResNet-101 has 100 convolutional layers with a 1,000-FC layer.
The average pooling layer and the convolutional layers are removed, and the convolutional layer is used to compute the feature maps.
A layer applied to the last convolutional block generates the score maps.
A sibling convolutional layer is also applied to calculate the bounding box regression.
On the PASCAL VOC 2007, it achieves an 83.6\% mAP with the 101-layer ResNet.
It suggests the same accuracy as the Faster R-CNN but achieves 20 times the speed of its Faster R-CNN counterpart.
Thus, R-FCN introduces two advantages over its predecessors: First, CNN is faster than FC layers. Second, the network becomes scale-invariant since there is no FC to restrict the input image size.

FPN (Feature Pyramid Networks for Object Detection) \cite{lin2017feature}: This method was designed to address an issue with Faster R-CNN. Faster R-CNN was generally made to address the scale-invariance problem introduced by Fast R-CNN. Faster R-CNN takes an input image and resizes it accordingly, meaning that the network has to run on the image several times with different box sizes, making it slow. The feature pyramid network (FPN) deals with these different scales while maintaining the speed. The FPN is an extension of Faster R-CNN in the same manner that R-FCN is an extension of Faster R-CNN. Having a robust scale invariance is important for object detection since the network should be able to recognize an object at any distance from the camera. Faster R-CNN aimed to tackle this issue by creating anchor boxes. This proved time-consuming since the anchor boxes had to be applied to each RoI. The FPN, however, creates multiple feature maps that aim to represent the image at different scales. Hence, the feature map in RPN is replaced by the FPN, removing the necessity of having multi-scale anchor boxes. The regression and classification are applied across these multiple feature maps. The FPN takes in an input image and outputs multiple feature maps representing smaller height and width but deeper channels known as the bottom-up pathway.
The feature maps generated by the FPN goes through a 1x1 convolutional layer with a depth of 256.
The lateral connection is then applied, which adds the feature elements to the upsampled version of the feature map.
Faster R-CNN runs on each scale, and predictions for each scale are generated.
FPN comprises two paths: the bottom-up that uses ResNet and the top-down.
In the bottom-up approach, CNN is applied to extract features.
On the top-down pathway, the FPN constructs a higher resolution layer, but the object locations are no longer accurate because of the down and upsampling. Therefore, FPN adds a lateral connection between the constructed layers to increase the probability of predicting locations. This method runs at 5 fps, as benchmarked by the previous methodology with a state-of-the-art result on the COCO 2016 dataset.
Images have objects with different scales, making it challenging to detect them.
When using several anchor boxes to detect objects with different scales, the ratio seems to be memory- and time-consuming. FPN seems to push the accuracy boundaries by introducing a pyramid of feature maps to detect objects of different sizes and scales in an image. It is important to highlight that FPN is a feature detector and not an object detector. Therefore, FPN has to be used with an object detector in its RoI.

Mask R-CNN \cite{he2017mask}:
This method extends Faster R-CNN by adding another layer to predict the object mask in parallel with the existing bounding box layer.
This is a framework that enables instance segmentation on a state-of-the-art level.
The mask branch added to the Faster R-CNN is a small FCN applied to each RoI, which predicts on a pixel-to-pixel basis.
Briefly, the Faster R-CNN has two stages: the RPN and the Fast R-CNN combined.
The Mask R-CNN adopts the same notion as an identical first stage of RPN, and in the second stage, it outputs a mask for each RoI in parallel to the predicting class and box.
The branch added to the second layer is an FCN on top of a CNN feature map.
The ROI poolings lead to misalignment; therefore, the RoIAlign layer is proposed to preserve the pixel-level alignments.
The main method Mask R-CNN introduces is the RoIAlign, which preserves the pixel–spatial correspondences and replaces the quantization from the RoI pooling with bilinear interpolation.
The state-of-the-art results are achieved by ResNeXt101-FPN in the Coco dataset.
The additional mask branch added introduces minor computational additions.
Mask R-CNN is a very promising instance segmentation method that is very flexible and efficient for instance-level segmentation. However, as with the original Faster R-CNN, this architecture struggles with smaller-sized objects, mainly because of the feature maps’ coarseness.

Other image classifications and object detections include but are not limited to NOC, Bayes, MR-CNN and S-CNN, HyperNet, ION, MSGR, StuffNet, OHEM, SDP+CRC, SubCNN, GBD-Net, PLANET, NIN, GoogLeNet, VGGNet, ResNet, DenseNet, RetinaNet, ResNet, Corner Net, Inception, Hourglass, Dilated Residual Networks, Xception, VGG, DetNet, Inception, Dual Path Networks (DPN), FishNet, ResNeXt, and GLoRe \cite{zhao2019object, liu2020deep,zhang2013object}.

For microcirculation analysis, we conclude that deep convolutional neural networks have lifted much of the burden for feature engineering, which was the main focus in the pre-D-CNN era, and changed the focus to designing more accurate and efficient network architecture. Despite the great successes, all methods suffer from the intense labor of creating the bounding boxes. All ‘newer’ methods need exponentially more RAM and GPU in exchange for increased accuracy.

Furthermore, detecting small-size objects and localizing these objects remains a challenge.
Using the stated architectures still requires an experienced machine learning engineer to select the appropriate parameters of the algorithms in order to learn the patterns of the small-sized objects.
Several solutions have been suggested by the literature, including multi-task learning (Stuffnet) \cite{brahmbhatt2017stuffnet}, multi-scale representation (IONet) \cite{bell2016inside}, and context modeling (HyperNet) \cite{kong2016hypernet}. On the other hand, methods have been proposed to deal with large data imbalances between the objects and the background, such as the online mining algorithms (OHEM) \cite{shrivastava2016training}.
For microcirculation analysis, we believe that a Region Proposal Based Framework achieves better microcirculation data accuracy overall.

\subsection{Unified Based Framework}

You Only Look Once (YOLO) \cite{redmon2016you}: YOLO is a Unified Based Framework for object detection suggested by Redmon et al \cite{redmon2016you}. The most significant difference between this architecture and the methods in the Region Proposal Based Framework is the ability to track objects in real time. As mentioned earlier, Fast R-CNN (Regions with Convolutional Neural Networks) proposes 2,000 regions to be predicted, while YOLO takes that down to 100 regions.
On a Titan X GPU, YOLO can classify up to 45 frames per second compared to Fast R-CNN at 0.5 frames per second. YOLO takes a 224x224 image as an input and divides the image into several grids. It then classifies each object within that grid by giving it two scores: what class it belongs to and confidence percentage. The classification is done by a 24-convolutional layer with a two-FC layer. According to the tests, YOLO was ineffective at localization, and had low accuracy with comparison to R-CNN.
Despite the high speed of YOLO, the low accuracy makes it an unsuitable choice for microcirculation analysis.

YOLOv2 \cite{redmon2017yolo9000}: YOLOv2 addresses the precision issues brought by YOLOv1. It first replaces the CNN classifier with DarkNet19 instead of GoogLeNet.
DarkNet19 is a simpler classifier utilizing 19 convolutional layers followed by five max-pooling layers, allowing for faster performance on the same dataset. It also removes the FC layer for prediction and uses the anchor boxes method instead, increasing the recall accuracy by 7\%. Batch normalization is added between each convolutional layer, increasing the mAP by 2\%. Furthermore, it increases the image input from 224x224 to 448x448, which increases the mAP by an additional 4\%.  In faster R-CNN, the size of the anchor boxes is defined beforehand, and YOLOv2 utilizes k-means clustering on the training set to find the right aspect ratio of the anchor boxes to use, increasing its accuracy by a further 5\%.

YOLOv3 \cite{redmon2018yolov3}: This is an improved version over the YOLOv2 that increases overall accuracy with multi-scale labeling of small objects. YOLOv3 uses three separate feature maps to predict the ROIs. It also uses DarkNet53 with independent logistic classifiers, allowing it to detect multi-overlapping objects in the image. With these changes, YOLOv3 is suited to detect smaller objects within the grid, but it performs worse with medium to larger objects.

Single Shot Multibox Detector (SSD) \cite{hu2019introductory}: This improved the detection precision of a one-stage detector by implementing multi-reference and multi-resolution detection techniques. SSD detected objects of different sizes and scales across the network instead of just applying detection on the last layer. SSD maintains the speed of YOLO but has higher accuracy on the same standardized sets used to benchmark YOLO. SSD uses VGG16 as its backbone for image classification.

RetinaNet \cite{lin2017focal}: This introduces focal loss, which increases the prediction accuracy on small and medium objects compared to the previously mentioned detectors. In an image, the object of interest is relatively smaller than the background image. Therefore, the number of background images creates a class imbalance. The focal loss function aims to increase the weight of the minority class while reducing the weight associated with the majority class. RetinaNet archives comparable accuracies with the region proposal based framework at the expense of speed.

CornerNet  \cite{law2018cornernet} challenges the use of anchor boxes by stating that they create the data imbalance issue in the first place. It also states that anchor boxes create unnecessary parameters that have to be tuned, which slows down the training and prediction time. Instead, CornerNet uses key points in a bounding box with a single convolutional neural network. It achieves the highest accuracy when compared to the standard benchmark dataset; however, it is slower than YOLO.

When examining the architectures in the Unified Framework, we generally notice a trade-off between speed and accuracy. With the above methods, as accuracy increased, the time for detection also increased.
In microcirculation analysis, having an accurate method is more important than a fast method.
Moreover, the difference in time analysis between the unified framework and the region proposed framework in microcirculation image analysis can boil down a few seconds.
Therefore, we recommend the use of region proposed framework for microcirculation analysis.

\subsection{Upscaling Images using Deep Neural Networks}

From our research, the microscope videos have very low resolution.
Upscaling the image might help the researcher annotate the data better.
The upscaling process involves improving an image’s details by increasing the dimensions and interpolating those extra pixels using a mathematical method.
These mathematical methods include Enhanced Deep Super-Resolution Network (EDSR) \cite{lim2017enhanced},  Efficient Sub-pixel Convolutional Neural Network (ESPCN) (\cite{shi2016real}, Fast Super-Resolution Convolutional Neural Network (FSRCNN) \cite{dong2016accelerating} and Laplacian Pyramid Super-Resolution Network (LapSRN) \cite{lai2018fast}.
EDSR employs an architecture similar to ResNet without the batch normalization layer and the ReLU activation layer after the residual block. This architecture can be used to create a scale factor of 2.  ESPCN extracts the feature maps and applies the upscaling at the end of the network. Like ESPCN, FSRCNN applies upscaling at the end of the network with a smaller filter size. LapSRN is based on the Laplacian pyramids concept, upscaling gradually through the network.

\section{Traditional Computer Vision Object Detection Techniques}
\label{traditional}

In this part, we present computer vision object detection techniques that can be used for microcirculation analysis.
These presented methods do not use neural networks for classification.
Such techniques are also known as feature descriptors; they were gaining momentum from the early 1990s until the rise of deep learning in 2012 \cite{liu2020deep}. Although feature descriptors have fallen out of favor compared to deep learning with the benchmark datasets, they are still very relevant for microcirculation analysis. Their computational power and simplicity make these algorithms easier to implement on low-powered or battery-powered devices in hospitals.

Computer vision techniques aim to locate the image of interest from the background by distinguishing between edges, colors, textures, corners, and other image properties.
Such traditional computer vision techniques needs the values coded beforehand, which were found via trial-and-error methods and domain expertise.
% Deep learning, on the contrary, applies non-linear methods to detect this automatically.
Below are three computer vision techniques detection methods that can used for microcirculation analysis.

Template matching-based object detection \cite{dufour2002template} methods consist of two steps.
The first step is the template generation step, in which a template is generated by an expert based on the training set; the second step involves matching new data with that template-based image.
A similar measure is then applied to detect similarities between these images. Statistical methods like the sum of absolute differences or Euclidean distances can quantify the similarities between the template and test data.
The template matching detection stage can be further categorized into methods: Rigid Template Matching (RTM) and Deformable Template Matching (DTM).
Further modifications for the stated template methods include the Scale-Invariant Feature Transform (SIFT), the Speeded-Up Robust Features (SURF), and the Binary Robust Independent Elementary Features (BRIEF).

The main disadvantage of RTM is that it is sensitive to slight changes in viewpoint, shadows, and other challenges, as was stated earlier, while DTM needs a lot of geometrical engineering in the template beforehand. Moreover, these templates require two independent parameters to be tuned, the template to be generated from the training set, and the most suitable method for measuring similarities to be selected. This makes this approach time-consuming for the case of microcirculation analysis.

Another set of methods involves knowledge-based object detection \cite{greig2003knowledge}. These can be further divided into geometric knowledge and context knowledge. A priori knowledge of the shape is encoded into the geometric knowledge methods. However, this is extremely difficult with capillaries since the shapes are irregular. Context knowledge encodes the spatial relationship between the object and the background around how the neighboring pixels interact. Again, due to the different shades of the skin and the blood, this method is not preferred for microcirculation analysis.

Object detection based on object-based image analysis (OBIA) \cite{hossain2019segmentation} is the most promising for microcirculation analysis and comprises two parts: the image segmentation part and the object classification part.
OBIA aims to group similar pixels together based on statistical methods.
In the case of microcirculation analysis, we would like to highlight the pixels of capillaries and separate them from the background.
Promising methods under the OBIA techniques are background subtraction methods, geometric transformation methods, and image thresholding techniques.
These methods can be used in combination with each other or independently to detect capillaries in an image.

\subsection{Background Subtraction Methods}

Background subtraction is a step in image preprocessing, where the goal is to remove the background and keep the object of interest \cite{benezeth2010comparative}.
The three methods stated next can take in an image of microcirculation and attempt to calculate an approximation of the location of the capillaries.

\begin{itemize}

    \item 
    Mixture of Gaussian Method: this method uses Gaussian mixture-based background/foreground segmentation \cite{kaewtrakulpong2002improved}. It takes in a pixel with a K Gaussian distribution and attempts to model the background. This method is based on using the L-recent window version after the sufficient statistics equation is calculated,
    
    \item 
    Improved Mixture of Gaussian Method: this method is also based on a Gaussian mixture-based background/foreground segmentation but uses recursive equations to update the parameters \cite{zivkovic2004improved}. The previous method selects the background based on K Gaussian distribution, while this method uses an adaptive density estimation \cite{zivkovic2006efficient},

    \item 
    Statistical background image estimation: this method uses Bayesian segmentation with Kalman filters and Gale-Shapley \cite{li2004statistical} matching to approximate the background image.

\end{itemize}

Marcomini et al. \cite{marcomini2018comparison} compare the performances of all three abovementioned methods using accuracy rate, precision rate, and processing time and conclude that the improved mixture of Gaussian methods had the best performance in their experimental dataset. This has also been shown to be the best method among the three for background selection in the CapillaryNet paper \cite{abdou2022capillarynet}.

\subsection{Image Thresholding Techniques}

In its simplest form, thresholding involves changing a pixel value if it is above or below a certain value \cite{sezgin2004survey}. This value or threshold can be determined by several methods, and the change of value can also be calculated by different methods. In microcirculation, this can be beneficial for determining the set of values that represent the capillary and the other that represents the background. 
Listed below are five thresholding techniques that can be used for microcirculation analysis.
%In the next section, we will look at different types of thresholding to determine the change in value and the magnitude of change.

\begin{itemize}

\item Binary Threshold: this method takes in two values—the threshold value and the value to be given if the value is higher than the threshold value. The values under the threshold value will be set to zero,

\item Truncating Threshold: similar to the above method, it takes in two values. However, anything lower than the threshold value remains the same, while anything higher gets the assigned value,

\item Zero Threshold: anything lower than the threshold value becomes zero, while anything higher stays the same.

\end{itemize}

The above methods are fairly simple, and these values are determined by the user. However, they are not the optimal method if different parts of the same image have different illumination. The object of interest may have a higher or lower value depending on the light; therefore, the next two methods were designed to deal with this issue.

\begin{itemize}

\item Adaptive Thresholding \cite{bradley2007adaptive}: thresholding is applied locally on some pixels rather than globally on the whole image. Thresholding can be calculated in two ways: mean of an area or weighted sum where the weights are decided by a Gaussian window. The size of the window is decided by the user. This way, every window-sized part of the image gets a threshold applied to it based on a calculation,

\item OTSU Binarization \cite{yousefi2011image}: this method is optimal for images with two peaks in their histogram. It finds a value between the two peaks in a histogram where the variance is minimal for both classes and applies thresholding based on that.

\end{itemize}

\subsection{Edges and Lines}

There are several methods for detecting edges and outlines of the capillaries. Below, we list those most relevant to microcirculation detection.

\begin{itemize}

    \item 
    Contours \cite{bradski2008learning}: this involves drawing a line joining the pixels with the same color or intensity. In our case, this can help highlight the outline of a capillary. This method has the highest accuracy when thresholding is applied beforehand, so more pixels have similar values. In the below method, we use a marching square algorithm, which linearly interpolates the pixel values to find the algorithm output \cite{bradski2008learning},
        
    \item 
    The Canny edge detector \cite{mcilhagga2011canny} can detect and quantize the capillary area. This is a multistage detector that uses a Gaussian derivative to compute the gradients. The Gaussian attempts to reduce the noise in the image, and the curves are detected by selecting the maximum pixel value,
    
    \item 
    Skeletonization \cite{sherstyuk1999kernel} is a method used to find the central pixels within the border image to get the object topology. This method iterates over the image several times, starting from the border of the object and moving toward the center until it terminates.
    
\end{itemize}

\subsection{Histogram Equalization}

An image can be enhanced using histogram equalization methods.
Histogram equalization \cite{kaur2011survey} can be done using three methods: standard equalization, contrast stretching or adaptive equalization.
\begin{itemize}

    \item In a standard equalization, the most frequent value is spread out to roughly have a linear cumulative distribution graph,
    \item In contrast stretching, the image pixels are rescaled to include all values between the 2nd and 98th percentile,
    \item With adaptive equalization, changes in pixels occur locally based on a window size rather than the whole image. 

\end{itemize}

\subsection{Image Denoising}

Image can be enhanced by reducing the noise.
This is called image denoising \cite{motwani2004survey}.
There are several ways to denoise an image; total variation filters, bilateral, wavelet denoising filters, and non-local means denoising algorithm.
\begin{itemize}

    \item 
    The total variation filter uses the L1 norm gradient to remove noise from the image,
    
    \item 
    The bilateral filter averages the pixels based on the weighted average of the window used by the user,
    
    \item 
    The wavelet denoising filter represents the image as a wave and analyzes the wavelet coefficients. The wavelet coefficients that are under a certain threshold are represented as noise and are removed, and
    \item 
    A Non-Local Means Denoising algorithm estimates the value of a pixel based on similar patches from other similar areas in the image. This method can be applied either by spatial Gaussian weighting or uniform spatial weighting.
    
\end{itemize}

\section{State-of-the-Art Microcirculation Image analysis Techniques} 
\label{state_of_the_art}

In this section, we present the methods used by other researchers to develop their microcirculation analysis systems. The following methods utilize computer vision techniques to segment and, in some cases, quantify the capillaries.

Dobble et al. \cite{dobbe2008measurement}  use a frame averaging method to remove the plasma and white blood cell gaps within the capillary before using an algorithm to detect capillaries. Using frame averaging can lead to a lower overall density calculation since capillaries with a majority of gaps or not enough blood flow will be disregarded. Furthermore, Dobble et al. \cite{dobbe2008measurement} remove capillaries that are out of focus since they consider it to add noise to the frame averaging method. From our experiments with handheld microscopy, the nature of the rounded lens may lead to 40\% out-of-focus images on both edges of the video. It is very challenging to have a fully focused video the whole time, and some parts can always be out of focus. Therefore, this will significantly reduce the capillary density values further.

Hilty et al. \cite{hilty2019microtools} have a similar flow to Dobble et al. \cite{dobbe2008measurement} with minor changes.
Hilty et al. \cite{hilty2019microtools} detect capillaries by generating a mean image across all frames then passing the resulting image to two pipelines: first, classifying vessels of 20–30 µm in diameter as capillaries and second, classifying any vessels of up to 400µm in diameter as venules.
The capillaries are then passed to a modified curvature-based region detection algorithm \cite{deng2007principal} and is equalized using an adaptive histogram.  The result is a vessel map that contains centerlines across structures that are 20–30µm wide. As the authors of the curvature-based region detection algorithm state \cite{deng2007principal}, this type of detection is unintelligent and can lead to the detection of artifacts such as hair or stains of similar sizes. Furthermore, due to the challenges with the skin profile stated above, the mean of the images across the whole video is not always the best representation value since different parts of the video might have different lighting or capillaries can be out of optimal focus. Moreover, videos with slight motion will have to be completely disregarded since the central line is calculated across all frames instead of per frame.

Similarly to Dobble et al. \cite{dobbe2008measurement}, Bezemer et al. \cite{bezemer2011rapid}  improve the method by using 2D cross-correlation to fill up the blood flow gaps that plasma and white blood cells cause. This is a better method since the number of frames to be disregarded is reduced. However, 2D cross-correlation assumes a uniform blood flow and does not consider the dynamic change of flow between the gaps, which can inherently decrease the prediction accuracy.

Tam et al. \cite{tam2010noninvasive}  detect capillaries through a semi-automated method that requires the user to select points on the image. The algorithm then decides if there is a capillary present. Since this method relies on the user to select the capillaries, it cannot be used in a clinical environment due to the time of analysis of a microscopy video.

Geyman et al. \cite{geyman2017peripapillary} take a more manual approach by first using software to click away from the major blood vessels and then applying hardcoded calculations to detect the total number of capillaries based on the number of pixels in the region of interest. This is a manual approach and is highly susceptible to observer variations across different datasets.

Demir et al. \cite{demir2012automated} use a contrast limited adaptive histogram equalization (CLAHE) method \cite{reza2004realization} with a median filter and an adjustable threshold to detect capillaries on the weighted mean of five consecutive frames. However, this method needs to be adjusted depending on the illumination on the video and the skin thickness. This introduces a manual job where the user has to find the right combination of values for different videos or the same video with different illumination.

Cheng et al.~\cite{cheng2015reproducible} apply an image enhancement step followed by the manual highlighting of the capillaries by the user. The image enhancement process darkens the capillaries and increases the background brightness using a best-fit histogram method. Using their system, the user can further increase the contrast and smoothen the images manually to increase the differentiation of the capillaries from the background. Macros of this modification can then be generated and applied to all future captured microscopy videos. However, this macro generation assumes that the videos will be captured with the same brightness and thickness of the skin. Moreover, the image used is in grayscale; therefore, if there are any artifacts, they can be mistaken for capillaries.

Tama et al.~\cite{tama2015nailfold} use binarization followed by skeleton extraction and segmentation to quantify the capillaries. The binarization is applied to the green channel since they assume it has the highest contrast between the capillaries and the background. They use the top-hat transform method to reduce uneven illumination, followed by Wiener filtering to remove noisy pixels, and then the Gaussian smoothing method to smoothen the image. The OTSU thresholding method is then applied to segment the capillaries from the background. This method relies on the user finding a reference frame from the video that has the highest contrast.

The work described next uses ML techniques to segment and, in some cases, quantify the capillaries.

Prentašic et al.~\cite{prentavsic2016segmentation} used a custom neural network to segment the foveal microvasculature. Their neural network consists of three CNN blocks coupled with max-pooling along with a dropout layer followed by two dense layers. Their neural network was trained in 30 hours, and the segmentation took approximately two minutes per single image, with an accuracy of 83\%. The time taken and the high-end hardware used to analyze a single image make it unsuitable for clinical use since the users would like the results instantly.

Dai et al.~\cite{dai2020exploring} used a custom neural network similar to
Prentasic et al. for segmentation. However, Dai et al. used five CNN blocks instead of three. They used gamma correction and CLAHE for image enhancement. They reported an accuracy of 60.94\%, which is too low to be useful.

Nivedha et al.~\cite{nivedha2016classification} used the green channel of the image and used a non-linear support vector machine [143] to classify the capillaries. This method involved a manual step where the user had to crop the region of interest to improve the histogram equalization. Nivedha et al. performed different experiments comparing different denoising filters, such as Gaussian, Wiener, median, and adaptive median, and concluded that the Gaussian filter is the most suitable for their data. Furthermore, they compared different segmentation methods, including OTSU, K Means, and watershed, and concluded that the OTSU method was the most suitable for their data. The segmented images were then passed to an SVM, and they achieved an accuracy of 83.3\%.

Java et al.~\cite{javia2018machine} modify the ResNet18~\cite{he2016deep} to quantify capillaries and use the first ten layers of the architecture. The main limitation of the ResNet architecture is that images have to be resized to 224x224; however, most capillary images are less than 100x100. This means images have to be scaled up, which makes this method inefficient and uses more resources than needed. They reported an accuracy of 89.45\% on their data; however, ResNet 18 [79] has 11 million trainable parameters, and with such scaling up, training time can be up to several hours, and prediction time can be up to several minutes. This can make it slow and inefficient within a clinical setting. The training and test times were not reported in this paper.

Ye et al.~\cite{ye2020vivo}  utilized the concept of transfer learning and used the inception Single Shot Detector V2 (SSD-inception v2)~\cite{barba2020deep} to build their neural network. The SSD-inception v2 has high accuracy with reduced computational complexity, making it suitable for capillary detection [160]. On the other hand, they used a spatiotemporal diagram analysis for the flow velocity calculation. This method requires white blood cells or plasma gaps in order to detect the velocity accurately. Therefore, capillaries that lacked such characteristics had to be disregarded, reducing the overall efficiency of velocity classification. Furthermore, as stated by the paper's authors, the spatiotemporal method can be cumbersome and time-consuming. The accuracy was not reported in this paper.

Hariyani et al.~\cite{hariyani2020capnet} used a U-net architecture combined with a dual attention module~\cite{hu2018squeeze,woo2018cbam}. The images had to be resized to 256x256, and an accuracy of 64\% was reported. This accuracy is low for it to be used in a clinical setting.

The more accurate methods require semiautomatic analysis, whereas the more automatic methods are less accurate, making them unsuitable for a clinical setting.
Moreover, none of the existing works mentioned above reported using parallel frameworks to calculate capillary density. 
The authors of this paper represent a method that exceeds the accuracy and speed of the above mentioned papers.
CapillaryNet is fully automatic and able to classify microcirculation videos in $\sim$0.9s with 93\% accuracy \cite{abdou2022capillarynet} whilst CapillaryX presents a parallel frameworks to calculate capillary density \cite{abdou2022capillaryx}.

\section{Conclusion}

In this paper, we present the most promising deep learning and computer vision techniques that can automate microcirculation analysis, specifically the quantification of capillaries. Automating the quantification of capillary density might reveal important biomarkers to clinical personnel that might assist in helping critically ill patients with life-threatening diseases. With the automation algorithms, the analysis time can be reduced from minutes to several seconds and decrease interobserver variability.
We start by introducing the importance of analyzing microcirculation videos.
We then present the two prominent ways of automating the analysis of microcirculation videos: traditional computer vision techniques and deep learning techniques. 
We discuss the types of deep neural networks, then dive into details about the convolutional neural networks. Convolutional neural networks are the preferred method for analyzing images since they have the highest accuracy in image classification competitions. We present why convolutional neural networks are good at what they do and what challenges they can overcome. We then present the anatomy of a convolutional neural network by discussing the fully connected layer, the convolutional layer, and the pooling layer. Moreover, we present different types of convolutional neural networks that combine these three modules differently. Since convolutional neural networks can only classify images and cannot localize the regions of the capillaries, we present deep learning object detection techniques. The deep learning object detection techniques consist of two main frameworks: the unified based framework and the regional proposed framework. We present seven different algorithms on the regional proposed framework and six different algorithms on the unified based framework. We then discuss traditional computer vision object detection techniques, specifically, non-ML-based object detection methods like background subtraction methods, image thresholding techniques, edges and lines, and image enhancement techniques. Through the sections in this article, we have recommended the algorithms that can be used to develop an automated capillary detector and quantifier. Our contribution with this article is to assist researchers and developers with where to start looking if they are to develop an automated algorithm for capillary detection and quantification.

\printbibliography

\end{document}